\documentclass[letterpaper, 10 pt, conference]{ieeeconf}  %

\IEEEoverridecommandlockouts
\usepackage{amsmath} %
\usepackage{dblfloatfix}
\usepackage[pdftex]{graphicx}
\graphicspath{{figures/}}
\usepackage{hyperref}
\hypersetup{
    colorlinks=true,
    linkcolor=black,
    citecolor=black,
    filecolor=black,
    urlcolor=black,
}
\usepackage[T1]{fontenc}
\usepackage[utf8]{inputenc}
\usepackage{csquotes}
\usepackage[english]{babel}
\usepackage[export]{adjustbox}
\usepackage{caption}
\usepackage{subcaption}
\usepackage[colorinlistoftodos]{todonotes}
\usepackage{lipsum}
\usepackage{orcidlink}
\usepackage[capitalize]{cleveref}
\usepackage{balance}
\usepackage{multirow}
\usepackage{booktabs}
\usepackage{makecell}

\usepackage{placeins}%

\usepackage[style=ieee,
            doi=false,
            url=false,
            mincitenames=1,
            maxcitenames=1,
            minbibnames=6,
            maxbibnames=6,
            backend=biber]{biblatex}  %
\title{%
Prediction-Driven Motion Planning: Route Integration Strategies in Attention-Based Prediction Models
}

\author{
    Marlon Steiner\,\orcidlink{0009-0005-4025-9142}$^{1}$,
    Royden Wagner\,\orcidlink{0009-0000-7754-0345}$^{1}$,
    {\"{O}}mer {\c{S}}ahin Ta\c{s}\,\orcidlink{0000-0002-1249-260X}$^{2}$ and
    Christoph Stiller\,\orcidlink{0000-0003-4165-2075}$^{1}$%
    \thanks{
        $^{1}$Institute of Measurement and Control Systems, Karlsruhe Institute of Technology (KIT),
        Karlsruhe, Germany
        {\tt\small \{firstname.lastname\}@kit.edu}
    }%
    \thanks{
        $^{2}$FZI Research Center for Information Technology, Karlsruhe, Germany
        {\tt\small \{lastname\}@fzi.de}
    }%
}

\usepackage{textcomp}
\usepackage[absolute,showboxes]{textpos}
\usepackage{setspace}

\TPMargin{5pt}

\newcommand{\acceptancenotice}[2]{  %
    \begin{textblock}{0.82}(0.09,0.935)
        \setstretch{0.65} %
        \noindent{\footnotesize{\copyright #1 IEEE.
        Personal use of this material is permitted.
        Permission from IEEE must be obtained for all other uses, in any current or future media,
        including reprinting/republishing this material for advertising or promotional purposes,
        creating new collective works, for resale or redistribution to servers or lists,
        or reuse of any copyrighted component of this work in other works.

        \vspace{2mm}
        \noindent
        In Proceedings of the IEEE International Conference on Intelligent Transportation Systems (ITSC) -- Gold Coast, \textsc{Australia}, 18-21 November 2025.
        }}%
    \end{textblock}
}

\begin{document}
\maketitle

\acceptancenotice{2025}{https://doi.org/10.1109/ITSC57777.2023.10422605}

\thispagestyle{empty}
\pagestyle{empty}

\begin{abstract}
    Combining motion prediction and motion planning offers a promising framework for enhancing interactions between automated vehicles and other traffic participants.
    However, this introduces challenges in conditioning predictions on navigation goals and ensuring stable, kinematically feasible trajectories.
    Addressing the former challenge, this paper investigates the extension of attention-based motion prediction models with navigation information.
    By integrating the ego vehicle's intended route and goal pose into the model architecture, we bridge the gap between multi-agent motion prediction and goal-based motion planning.
    We propose and evaluate several architectural navigation integration strategies to our model on the nuPlan dataset.
    Our results demonstrate the potential of prediction-driven motion planning, highlighting how navigation information can enhance both prediction and planning tasks.
    Our implementation is at: \texttt{https://github.com/KIT-MRT/future-motion}.
\end{abstract} %
\section{Introduction}
\label{introduction}
In driving scenarios that involve interactions between traffic participants, accurately predicting others' future trajectories is essential for effective planning.
This is particularly important for autonomous vehicles, which must navigate complex environments while ensuring safety and efficiency.
\textit{Motion prediction} (prediction) and \textit{motion planning} (planning) are two key components of this process.
Prediction involves forecasting the future trajectories of surrounding agents based on their current and past states.
Planning, on the other hand, focuses on generating a safe and efficient trajectory for the ego vehicle to follow \cite{tas2022motion}.
Traditionally, these two tasks have been treated separately, which lacks modeling the bidirectional influence both tasks naturally have:
The predictions influence the planning, while the planning influences the reaction of the other agents.
However, recent advancements in deep learning and transformer architectures have led to promising research direction of \textit{integrated prediction and planning (IPP)} \cite{hagedorn2024ipp}.
Integrating these two tasks can improve the performance of both, as they can leverage shared information and enhancing the modeling of interactions between traffic agents.

\begin{figure}[!h]
    \begin{subfigure}{\columnwidth}
        \centering
        \includegraphics[width=0.85\linewidth, clip, trim=0cm 4cm 10cm 2cm]{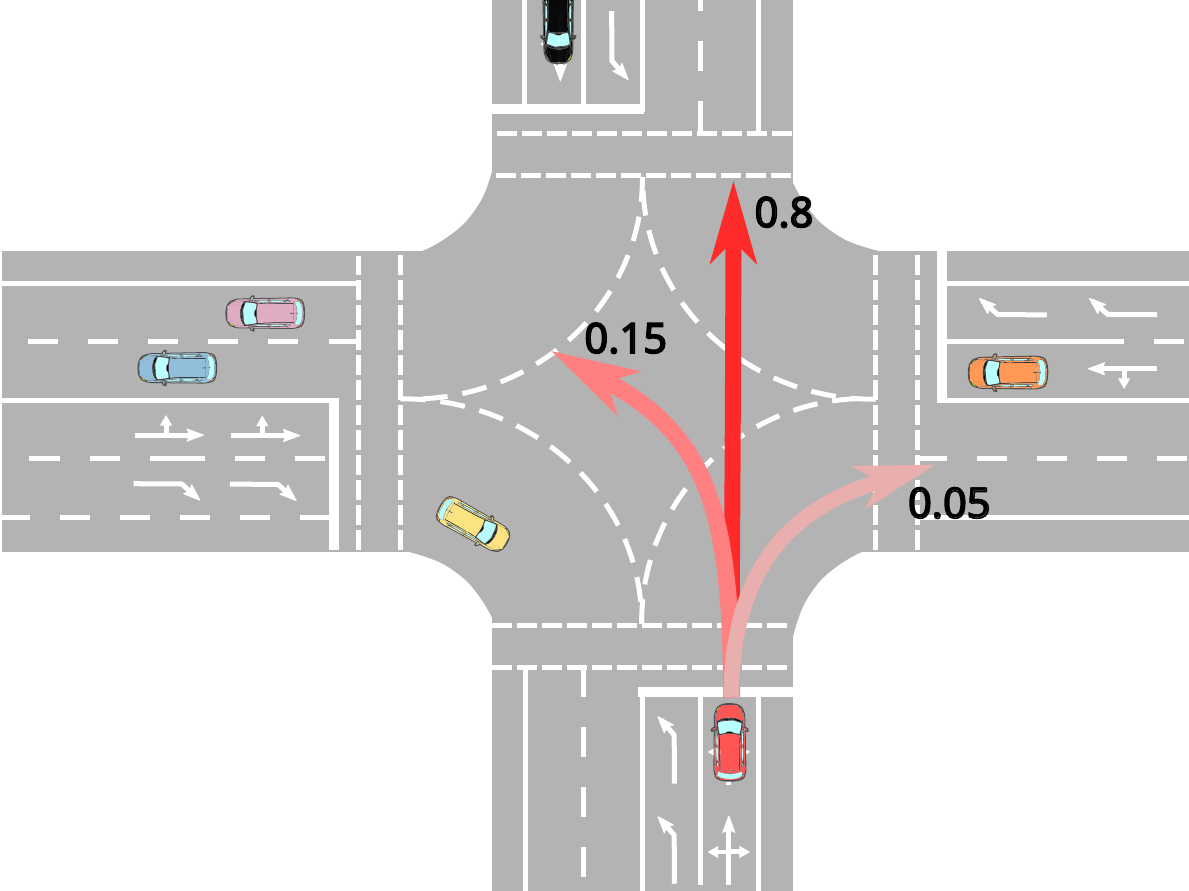}
        \caption{Ego vehicle's possible trajectories based on its past trajectory and context (e.g., map and other vehicles), without knowledge of the navigation goal.}
        \label{fig:motivation-a}
    \end{subfigure}

    \begin{subfigure}{\columnwidth}
        \centering
        \includegraphics[width=0.85\linewidth, clip, trim=0cm 4cm 10cm 2cm]{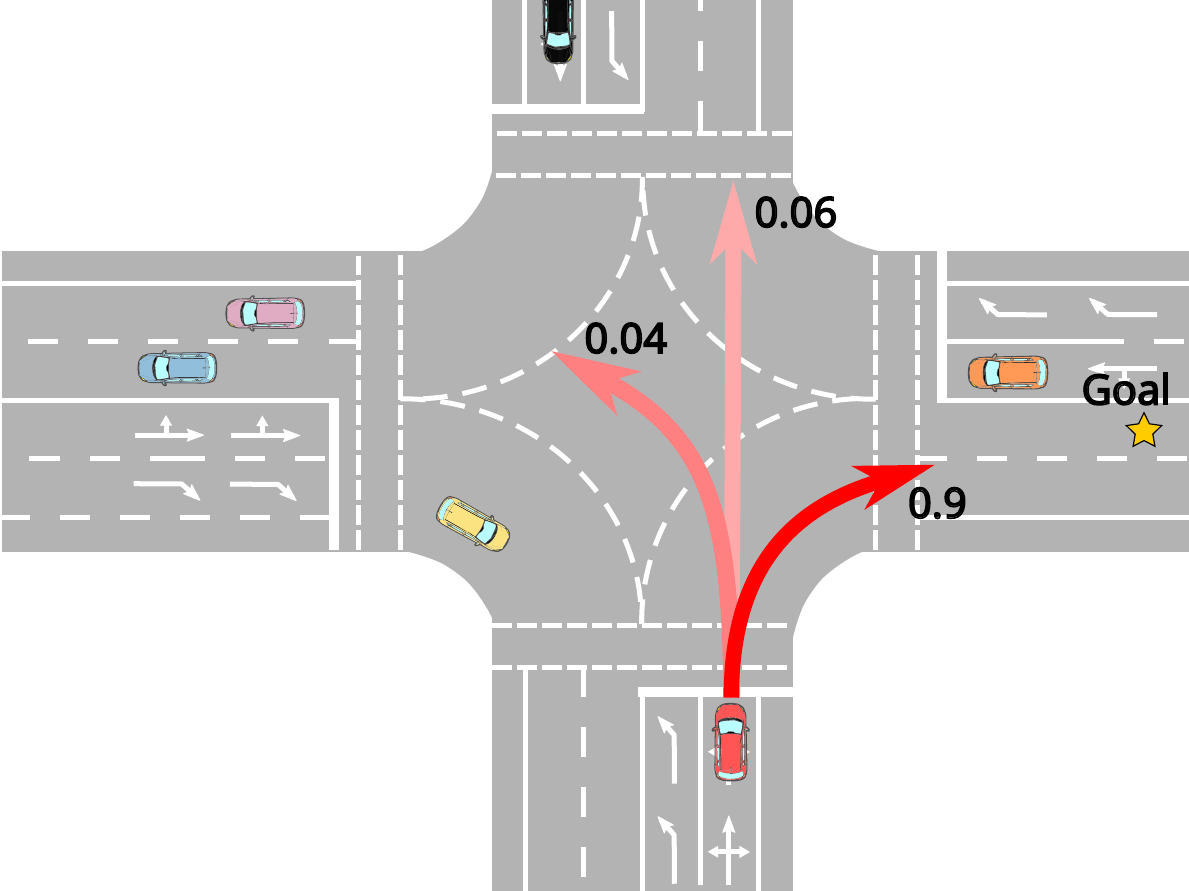}
        \caption{Ego vehicle's possible trajectories additionally conditioned on navigation information.}
        \label{fig:motivation-b}
    \end{subfigure}

    \caption{Motivation. The figures show three possible trajectories of the ego vehicle (red). Each trajectory is assigned a confidence, indicating how likely it is to be the actual trajectory.
        \cref{fig:motivation-a} does not include navigation information (i.e. goal and route), while \cref{fig:motivation-b} does.}
    \label{fig:motivation}
\end{figure}

This paper investigates how transformer- or attention-based prediction models can be extended to incorporate navigation information (goal and route), enabling them to function as effective motion planners.
\cref{fig:motivation} illustrates the motivation behind this approach.
The given example shows that conditioning the model on navigation information shifts the predicted probability of the multi-modal prediction.
This results in a higher confidence assigned to the trajectory that is most likely to reach the desired navigation goal.
We propose and evaluate several architectural strategies for integrating navigation information into our model, leveraging the nuPlan dataset for benchmarking.
We show that leveraging the given navigation information lead to improvements in prediction and planning.

In recent years, the need for reliable and standardized evaluation of motion planners has gained significant attention.
Various datasets and benchmarks have been introduced to address this, including \textit{nuPlan} \cite{caesar2022nuplan}, \textit{CARLA} \cite{dosovitskiy2017carla}, and the \textit{Waymo Sim Agents Challenge (WOSAC)} \cite{montali2023simagents}.
Among these, we focus on the nuPlan dataset due to its suitability for non-end-to-end and single-agent planning evaluations.

Our primary contribution lies in conducting a thorough evaluation of strategies for integrating navigation information, thereby laying the groundwork for advancing prediction-driven planning systems in future research.

\section{Related Work}
\label{sec:related_work}

\subsection{Motion Prediction}
Prediction methods can be broadly categorized into joint, conditional, and marginal approaches, each with unique strategies for modeling future agent interactions.
A common practice in motion prediction is to forecast multiple plausible future trajectories per agent to account for the inherent uncertainty in dynamic environments.
Joint prediction captures the distribution of future motions for multiple agents, often leveraging attention mechanisms to model interactions and decode consistent futures \cite{wagner2024scenemotion,steiner2024mapformer, zhang2023real, ngiam2022scene, seff2023motionlm, shi2024mtr++, mercat2020multi, girgis2022latent}.
Methods like autoregressive decoding \cite{seff2023motionlm} or denoising diffusion processes \cite{jiang2023motiondiffuser} enable temporally causal predictions but suffer from high inference latency due to sequential sampling.
Conditional approaches predict an agent's future motion conditioned on the behavior of other agents \cite{salzmann2020trajectron++, tolstaya2021identifying}, typically using iterative rollouts for scene-wide predictions, which can be computationally expensive in dense traffic scenarios.
Goal-conditioned methods \cite{ngiam2022scene} simplify the task by conditioning on destination points, thus linking prediction with planning objectives.
Marginal methods, which independently predict each agent's motion, often include auxiliary objectives to capture interactions \cite{sun2022m2i, luo2023jfp}, but this limits interaction modeling to combining independently generated forecasts.

\subsection{Motion Planning}
Planning methods can be broadly categorized into rule-based, learned, and end-to-end approaches, with recent advancements leveraging generative and language-based techniques.
Rule-based methods rely on predefined rules or optimization techniques to generate feasible trajectories.
One work \cite{dauner2023pdm} combines rule-based strategies with a small MLP network to enhance adaptability.
Recent learning-based approaches \cite{cheng2024pluto, zhou2024behaviorgpt, wu2024smart} mainly rely on transformer architectures and demonstrate the potential of learned models in diverse traffic scenarios.
End-to-end methods \cite{li2025navigationguidedsparsescenerepresentation, yuan2024mamba, chen2024drivinggp} directly map sensor inputs to driving actions or trajectories.
Generative approaches, particularly diffusion-based methods, have gained traction for their ability to model diverse and realistic trajectories.
Several studies \cite{huang2024gendrive, hu2024mpwithgenmodel, liao2024diffusiondrive} highlight the effectiveness of generative models in producing high-quality motion plans.
Language-based methods incorporate natural language understanding into motion planning.
For example, one approach \cite{chen2024AsyncDriver} leverages a language model to interpret driving instructions and generate corresponding trajectories.

\subsection{Integrated Prediction and Planning}
Since both tasks are closely related and require similar contextual information, IPP approaches attract attention.
IPP can be categorized into three groups \cite{hagedorn2024ipp}:
\subsubsection{Sequential IPP}
Here, prediction and planning are treated as separate and sequential tasks, where either it is planned first \cite{hu2023planning} and predicted afterwards or the other way around \cite{song2020pip}.
However, influencing the other task is unidirectional.
\subsubsection{Undirected IPP}
In undirected IPP, the interaction between both modules is not clearly defined.
It can be achieved by a fully E2E model \cite{li2025navigationguidedsparsescenerepresentation} or a joint optimization where both tasks are performed in a single step, like in a joint prediction model \cite{wagner2024scenemotion}.
\subsubsection{Bidirectional IPP}
This approach can leverage autoregressive decoding methods \cite{guo2024rethinkingMPP, huang2023gameformer} or theoretically grounded techniques \cite{liu2024betopnet}.
Additionally, planning through simulation \cite{konstantinidis2025conditionalpredictionsimulation} is also well-suited for this category.

While planning and prediction are facing similar challenges, like generating reasonable trajectories, they differ in two main aspects \cite{hagedorn2024ipp}:
\begin{enumerate}
    \item The navigation goal is known in planning, but not in prediction:
          The planned trajectory must be conditioned on a navigation goal.
    \item In planning, the task is to find a single suitable trajectory, which must be stable in closed-loop:
          It must be kinematically feasible and result in safe and efficient driving behavior when fed to a downstream controller.
\end{enumerate}
In the following, the two aspects are referred to as \textit{difference (1)} and \textit{difference (2)}.

\section{Methodology}
\label{sec:methodology}

\begin{figure*}[t!]
    \vspace*{2mm}
    \includegraphics[width=\textwidth, clip, trim=0cm 1.5cm 0cm 1.5cm]{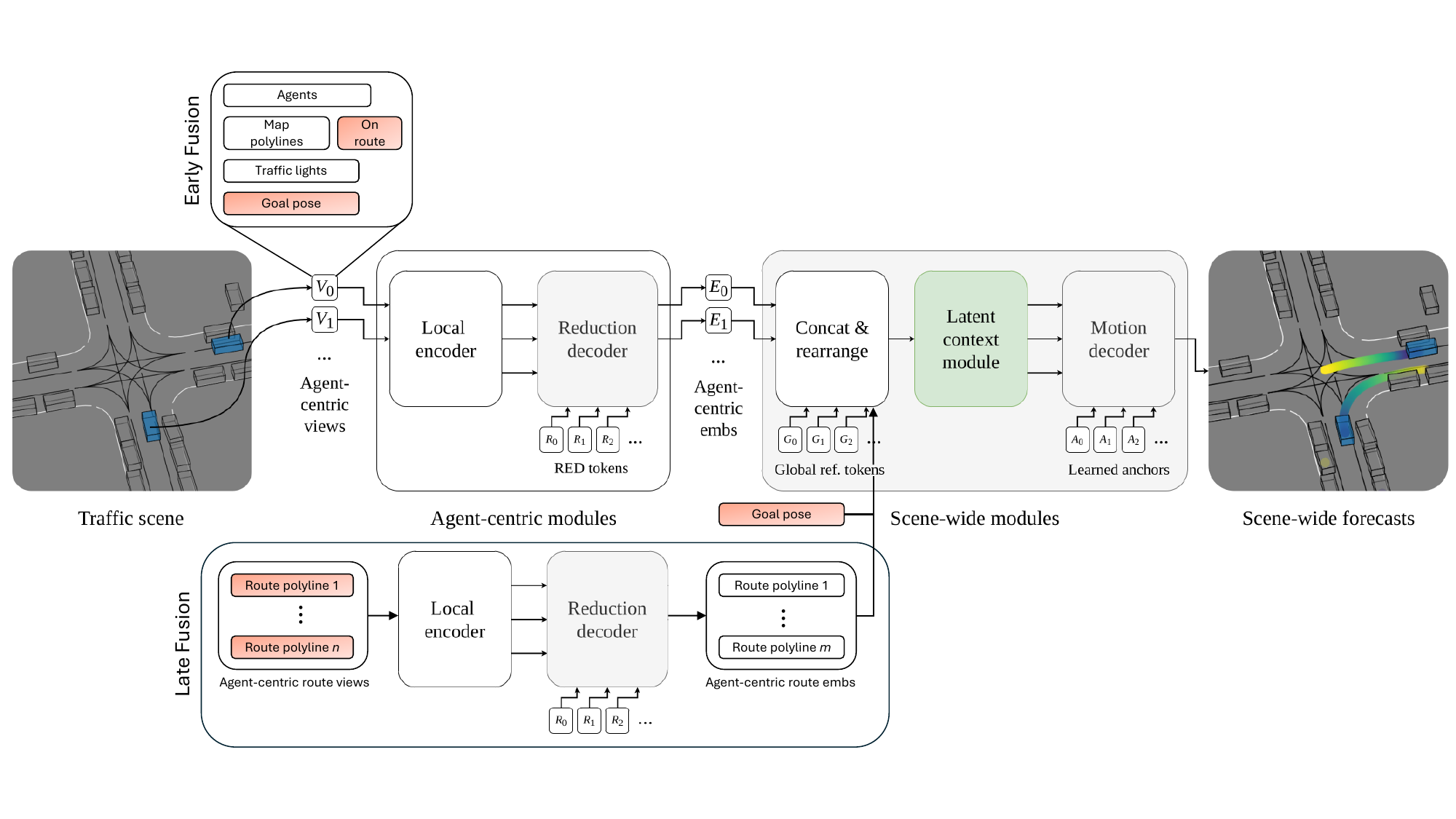}
    \caption{SceneMotion prediction model \cite{wagner2024scenemotion} and our extensions to fuse navigation information into the model (early fusion / late fusion).
        The inputs highlighted in red are additional information compared to the baseline model.}
    \label{fig:architecture}
\end{figure*}

Our goal of this study is to integrate navigation information into a prediction model, s.t. \textit{difference (1)} (unknown navigation goal, see section \ref*{sec:related_work}) is approached.
For now, \textit{difference (2)} (stability in closed-loop) will not be part of the study.

The nuPlan dataset provides two kinds of navigation information: First, information about which lanes belong to the route (route information) and second, the goal pose of a scenario (goal information).
Given the traffic rules, a lane is part of the route if a lane graph -- starting at the given lane -- can be constructed, which reaches the goal pose.
The goal pose is the final destination of the ego vehicle and will not be reached within the planning horizon.
To close the gap between a navigation unaware prediction model and a navigation aware planning model, we propose to integrate the navigation information into the prediction model.

Therefore, we compare a baseline model with three different integration strategies.
The baseline model is SceneMotion \cite{wagner2024scenemotion}. It predicts six different joint modes of up to eight agents in a scene, without knowing about the existence of an ego vehicle.
The model is based on a transformer architecture and uses a polyline representation \cite{zhang2023real} of the scene.
We provide an architectural overview of the model and the different navigation integration strategies in \cref{fig:architecture}.

\subsection{Navigation Integration Strategies}
In general, we differentiate between two different integration strategies: Early fusion and late fusion.
Additionally, we combine the two strategies with a navigation loss.
A detailed overview of the loss is given in \cref{sec:nav_loss}.

\textbf{Early Fusion w/o navigation loss:}
As a first approach, we fuse the given navigation information (route and goal information) in the early feature representation stage of the model.
While the baseline model gets different classes of map polylines (e.g. centerline, boundary, crosswalk, etc.) as input, we add a one-hot encoded vector to each polyline, which indicates whether the respective polyline belongs to the route or not.
We also add the goal pose as an additional token to the agent-centric view (V), allowing the model to incorporate the navigation goal directly into its feature representation.
This model will be referred to as \textit{SceneMotion-A1}.

\textbf{Early Fusion w/ navigation loss:}
In addition to the described early fusion strategy, we also add a navigation loss to the model.
This model will be referred to as \mbox{\textit{SceneMotion-A2}}.
Both, SceneMotion-A1 and SceneMotion-A2, introduce nearly the same number of parameters compared to the baseline and the additional required data is negligibly larger.

\textbf{Late Fusion w/ navigation loss:}
To analyze if the model benefits from fusing the navigation information at a later stage within the same model, we also implement a late fusion strategy.
Unlike the earlier stage, where different classes of polylines can be distinguished, the later stage employs an alternative approach to encode the navigation information.
Analogous to the agent-centric views (V), we encode the route information in a parallel branch.
The \mbox{\textit{reduction decoder} \cite{wagner2023redmotion}} reduces $n$ input route polylines to $m$ route embeddings, where $m<n$.
This is done to regularize the number of route embeddings compared to the number of agent-centric embeddings, comprising the scene context.
Together with an embedding for the goal information, these navigation embeddings are concatenated to the agent-centric embeddings (E) of the model.
Compared to the first two adaptions of the baseline, this model introduces slightly more parameters, which however, has a negligible impact on the training time.
This model will be referred to as \mbox{\textit{SceneMotion-A3}}.

\subsection{Navigation Loss}
\label{sec:nav_loss}
The navigation loss is a sensible extension to the model.
Contrary to the imitation-based loss in the baseline model, the navigation loss should not be based on the ground truth trajectories.
Instead, it should encourage the model to generate trajectories that are consistent with the route.
A key distinction between prediction and planning is crucial here:
While deviations from the ground truth negatively impact the prediction task, such trajectories can still be valid or even desirable in the context of planning.
Therefore, a loss based on a ground truth or goal point is not suitable for planning.
We propose a loss that just accounts for the lateral distance to the route, s.t. a predicted trajectory will follow the route, regardless of its progress.

The loss consists of two parts:
First, we calculate the lateral distance $d_\mathrm{lat}$ of the last predicted point of the most probable trajectory to the route.
The route is defined as all centerlines of lanes that are part of the routing graph.
Second, we substitute $d_\mathrm{lat}$ as $x$ in the function $f(x, \alpha, c)$, which is defined as follows \cite{barron2019CVPR}:
\begin{equation*}
    f(x,\alpha,c) = \frac{\alpha-2}{\alpha}\left( \left(\frac{\frac{x}{c}^2}{|\alpha-2|} + 1\right)^{\frac{\alpha}{2}} - 1 \right).
\end{equation*}
The function $f(x, \alpha, c)$ can be utilized as a robust loss function, capable of interpolating between various classical robust loss functions by adjusting the parameter $\alpha$. The parameter $c$ serves as a scaling factor for the loss.
For our implementation, we set $\alpha = -5$ and $c = 3$.
With $\alpha = -5$, the loss function approximates the \textit{Welsch loss} (as $\alpha \to -\infty$), which asymptotically converges to 1.
This choice of parameters ensures a smooth and continuous loss function, facilitating stable training by mitigating the impact of outliers.

\section{Evaluation of Navigation Integration Strategies}
\label{sec:results_and_evaluation}

\begin{table*}
    \vspace*{1mm}
    \centering
    \caption{
        Val14 benchmark. We evaluate the original SceneMotion prediction model and modified versions of it, where navigation information is processed.
        We show prediction metrics for the baseline and several route integration strategies.
        The metrics are evaluated after 3\,s, 5\,s, and 8\,s of prediction.
        The \textit{Avg} is the average over all prediction horizons.
        Additionally, we show the average only for the ego vehicle (Ego (Avg)).
        Since this option only evaluates a single vehicle, the OR metric is not applicable.
        The best results are highlighted in \textbf{bold} and the second best are \underline{underlined}.
    }
    \label{tab:val14}
    \begin{tabular}{llccccc}
        \toprule
        Method & Horizon (s) & mAP\,$\uparrow$    & minADE\,$\downarrow$ & minFDE\,$\downarrow$ & MR\,$\downarrow$   & OR\,$\downarrow$   \\
        \midrule

        \multirow{5}{*}{SceneMotion \cite{wagner2024scenemotion}}
               & 3           & 0.4098             & 0.2937               & 0.5410               & 0.1582             & 0.2503             \\
               & 5           & 0.3750             & 0.4219               & 0.9535               & 0.1482             & 0.2718             \\
               & 8           & 0.3235             & 0.5898               & 1.7504               & 0.1614             & 0.2974             \\
               & Avg         & 0.3694             & 0.4351               & 1.0816               & 0.1559             & 0.2732             \\
               & Ego (Avg)   & 0.4819             & 0.3443               & 0.7307               & 0.0771             & --                 \\
        \midrule

        \multirow{5}{*}{\makecell[l]{SceneMotion w/ early route fusion                                                                    \\w/o navigation loss\\(SceneMotion-A1)}}
               & 3           & 0.4131             & 0.2871               & 0.5279               & 0.1567             & 0.2502             \\
               & 5           & 0.3783             & 0.4134               & 0.9395               & 0.1471             & 0.2722             \\
               & 8           & 0.3339             & 0.5780               & 1.6905               & 0.1522             & 0.2980             \\
               & Avg         & \textbf{0.3751}    & \textbf{0.4262}      & \underline{1.0526}   & \textbf{0.1520}    & 0.2735             \\
               & Ego (Avg)   & \textbf{0.4987}    & \textbf{0.3112}      & \textbf{0.6548}      & \textbf{0.0660}    & --                 \\
        \midrule

        \multirow{5}{*}{\makecell[l]{SceneMotion w/ early route fusion                                                                    \\and navigation loss\\(SceneMotion-A2)}}
               & 3           & 0.4039             & 0.2931               & 0.5432               & 0.1648             & 0.2497             \\
               & 5           & 0.3648             & 0.4211               & 0.9554               & 0.1542             & 0.2718             \\
               & 8           & 0.3158             & 0.5887               & 1.7274               & 0.1594             & 0.2972             \\
               & Avg         & 0.3615             & 0.4343               & 1.0753               & 0.1595             & \underline{0.2729} \\
               & Ego (Avg)   & 0.4860             & 0.3251               & 0.6806               & 0.0698             & --                 \\
        \midrule

        \multirow{5}{*}{\makecell[l]{SceneMotion w/ late route fusion                                                                     \\and navigation loss\\(SceneMotion-A3)}}
               & 3           & 0.4007             & 0.2879               & 0.5267               & 0.1560             & 0.2487             \\
               & 5           & 0.3648             & 0.4137               & 0.9374               & 0.1481             & 0.2707             \\
               & 8           & 0.3161             & 0.5787               & 1.6737               & 0.1540             & 0.2977             \\
               & Avg         & \underline{0.3605} & \underline{0.4267}   & \textbf{1.0459}      & \underline{0.1527} & \textbf{0.2724}    \\
               & Ego (Avg)   & \underline{0.4917} & \underline{0.3237}   & \underline{0.6625}   & \underline{0.0679} & --                 \\
        \bottomrule
    \end{tabular}
\end{table*}

In this section, we evaluate the proposed navigation integration strategies with the SceneMotion baseline on the nuPlan dataset and nuPlan open-loop challenge.
Since our scope is to address the integration of the navigation information into the prediction model, we do not evaluate in closed-loop.

\subsection{Dataset}
The model requires a selection of focus agents to predict their future trajectories.
Since the nuPlan dataset does not contain such a selection, we try to mimic the selection of the \textit{Waymo Open Motion Dataset} \cite{ettinger2021womd}.
They choose up to eight agents per scenario including at least two agents of each type (vehicle, pedestrian, cyclist) if available.
The selection is biased to include agents that do not follow a constant velocity model or straight paths.
Therefore, we calculate an interest score for every agent's trajectory in the scene based on \textit{total heading change}, \textit{lateral deviation}, \textit{acceleration}, and \textit{progress} and choose the agents of each type with the highest scores.

For the prediction evaluation in \cref{sec:eval_integration_strategies}, we use the validation split of the nuPlan dataset.
The open-loop planning in \cref{sec:eval_open_loop} is evaluated on the \textit{Val14} split \cite{dauner2023pdm}, which is close to the official test split from the nuPlan challenge 2023.

\begin{figure*}[!h]
    \vspace*{2mm}
    \begin{subfigure}[b]{0.24\textwidth}
        \includegraphics[width=\linewidth, clip, trim=0cm 1cm 0cm 1cm]{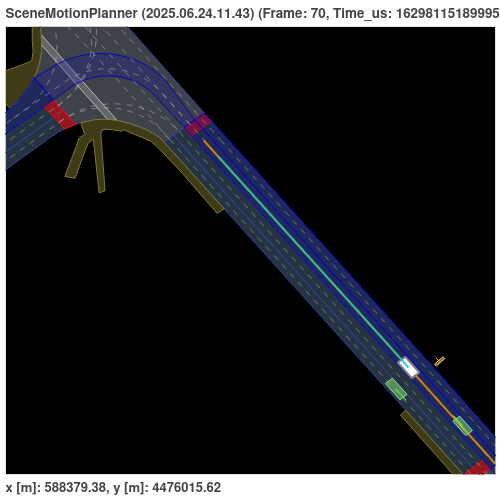}
        \caption{Frame 70}
        \label{fig:visual-evaluation-a}
    \end{subfigure}
    \hfill
    \begin{subfigure}[b]{0.24\textwidth}
        \includegraphics[width=\linewidth, clip, trim=0cm 1cm 0cm 1cm]{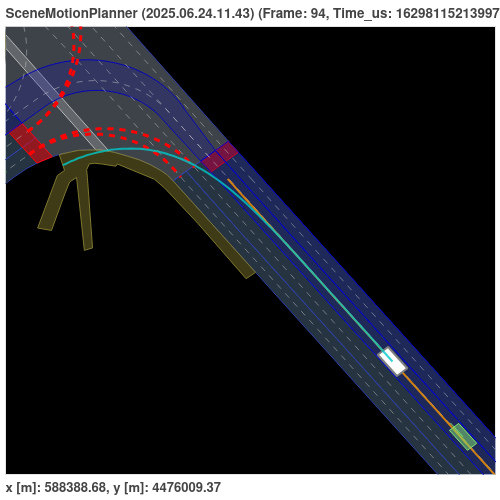}
        \caption{Frame 94}
        \label{fig:visual-evaluation-b}
    \end{subfigure}
    \hfill
    \begin{subfigure}[b]{0.24\textwidth}
        \includegraphics[width=\linewidth, clip, trim=0cm 1cm 0cm 1cm]{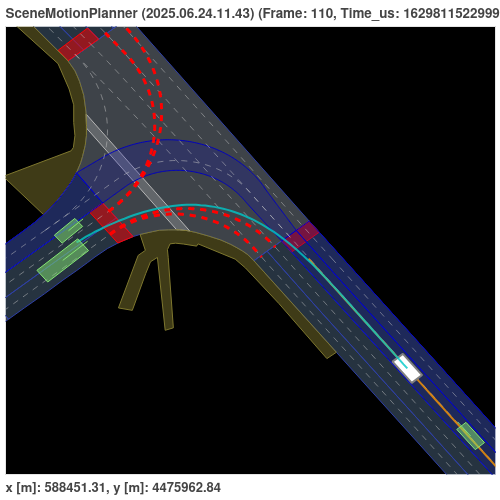}
        \caption{Frame 110}
        \label{fig:visual-evaluation-c}
    \end{subfigure}
    \hfill
    \begin{subfigure}[b]{0.24\textwidth}
        \includegraphics[width=\linewidth, clip, trim=0cm 1cm 0cm 1cm]{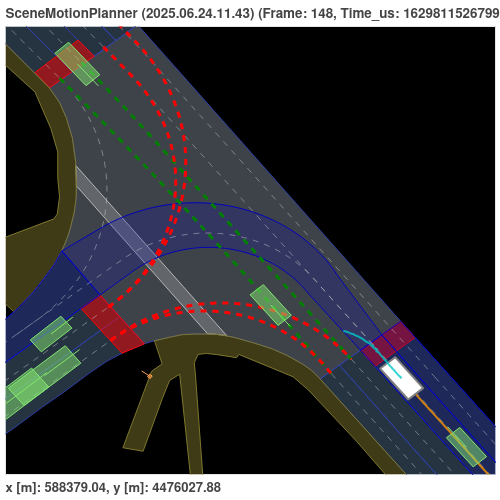}
        \caption{Frame 148}
        \label{fig:visual-evaluation-d}
    \end{subfigure}

    \vspace{1em} %

    \begin{subfigure}[b]{0.24\textwidth}
        \includegraphics[width=\linewidth, clip, trim=0cm 1cm 0cm 1cm]{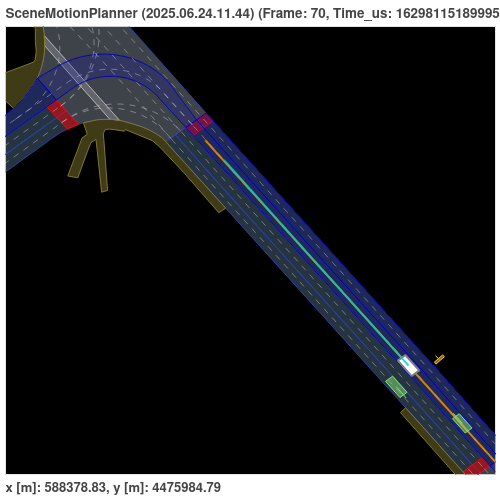}
        \caption{Frame 70}
        \label{fig:visual-evaluation-e}
    \end{subfigure}
    \hfill
    \begin{subfigure}[b]{0.24\textwidth}
        \includegraphics[width=\linewidth, clip, trim=0cm 1cm 0cm 1cm]{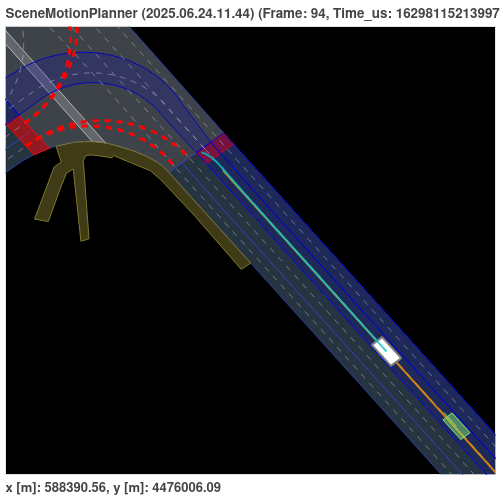}
        \caption{Frame 94}
        \label{fig:visual-evaluation-f}
    \end{subfigure}
    \hfill
    \begin{subfigure}[b]{0.24\textwidth}
        \includegraphics[width=\linewidth, clip, trim=0cm 1cm 0cm 1cm]{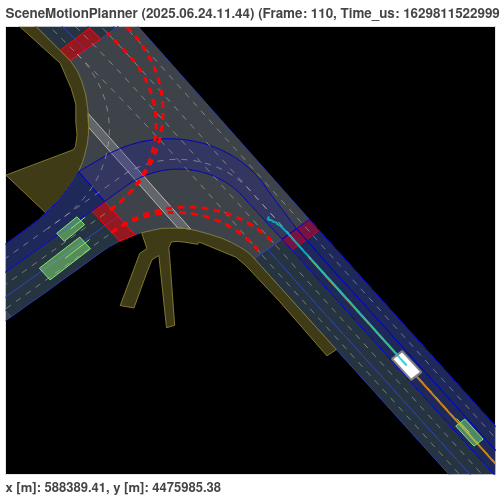}
        \caption{Frame 110}
        \label{fig:visual-evaluation-g}
    \end{subfigure}
    \hfill
    \begin{subfigure}[b]{0.24\textwidth}
        \includegraphics[width=\linewidth, clip, trim=0cm 1cm 0cm 1cm]{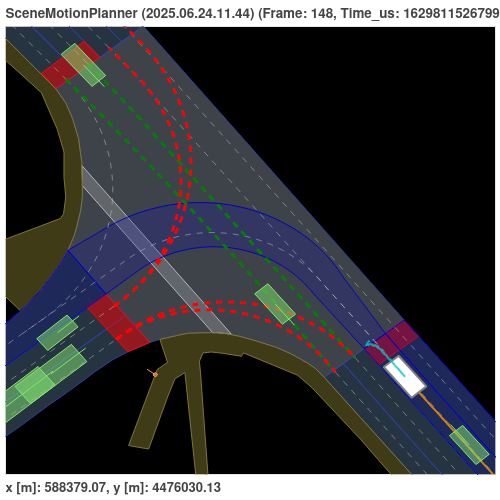}
        \caption{Frame 148}
        \label{fig:visual-evaluation-h}
    \end{subfigure}

    \caption{Left turn intersection scenario (nuPlan scenario token \texttt{bec10cb7c4985dbe}): A comparison of the SceneMotion baseline (top row) and the SceneMotion-A1 model (bottom row) at different times.
        The ground truth driven trajectory is shown in orange, while the planned trajectory are shown in blue.
        The ego vehicle is shown in white, while other agents are shown in green.
        Lanes visualized in blue indicate that they belong to the ego vehicle's route.}
    \label{fig:visual-evaluation}
\end{figure*}

\subsection{Training Details}
Compared to the original implementation, the model is configured with only 6 million parameters by halving the hidden dimension, resulting in a significantly smaller size.
While increasing the model size could improve performance, the current configuration is sufficient for our purposes.
We sample 112 scenes in a batch, with 8 focal (i.e., predicted) agents per scene.
We use Adam with weight decay \cite{loshchilov2017decoupled} as the optimizer and a step learning rate scheduler to halve the initial learning rate of $2 \times 10^{-4}$ every 20 epochs.
We train for 50 epochs using data distributed parallel (DDP) training on 4 A100 GPUs.
The training took approximately 48 hours.
Further implementation details can be found in the original SceneMotion paper \cite{wagner2024scenemotion}.

\subsection{Evaluation Metrics}
\subsubsection{Prediction}
We use the metrics from the \textit{Waymo Motion Prediction Challenge} to evaluate the models predictions: The mean average precision (mAP), the average displacement error (minADE), and the final displacement error (minFDE), the miss rate (MR), and the overlap rate (OR).
All metrics are computed using the minimum mode for $k = 6$ modes.
Accordingly, the metrics for the mode closest to the ground truth are measured.
All metrics are averaged over the 3 agent classes (vehicles, pedestrians, and cyclists).
We provide metrics for three prediction horizons of 3\,s, 5\,s, and 8\,s, and an additional average over these horizons.
We evaluate the given metrics for joint modes (i.e., best scene-wide mode).

\subsubsection{Open-loop Planning}
For the open-loop planning evaluation, we use the nuPlan open-loop score (OLS) (see \cref{tab:OLS}).
It is a weighted sum of the average displacement error (ADE), the final displacement error (FDE), the average heading error (AHE), the final heading error (FHE), and the miss rate (MR) over 3\,s, 5\,s, and 8\,s of prediction horizon.
The individual metrics closely align with the prediction metrics.

\subsubsection{Closed-loop Planning}
For completeness, the closed-loop planning evaluation on nuPlan is based on the following metrics:
At-fault collisions, drivable area compliance, driving direction compliance, making progress, time to collision, speed limit compliance, ego progress along the expert’s route ratio, and comfort.
We do not evaluate the closed-loop score in this work, but provide an overview of the relevant metrics to show that open-loop planning is closely related to prediction and differs from closed-loop planning.

\subsection{Influence of Integration Strategies on Prediction}
\label{sec:eval_integration_strategies}
In this section, we evaluate the influence of the navigation integration strategies on the prediction metrics.
The full results are shown in \cref{tab:val14}.
The baseline model (SceneMotion) achieves a mAP of 0.3694, a minADE of 0.4351, and a minFDE of 1.0816.
When only considering the prediction of the ego vehicle, these metrics are 0.4819, 0.3443, and 0.7307, respectively.
The evaluation of the proposed navigation integration strategies reveals several key insights.
Among the tested approaches, the best adaptation is achieved with early fusion of navigation information without the use of a navigation loss (SceneMotion-A1).
This configuration demonstrates significant improvements, particularly in the displacement error metrics.
For instance, both, the ego vehicle's minADE and minFDE improved by around $10\,\%$, the mAP improved by $3.5\,\%$ and the MR by $14.4\,\%$, in SceneMotion-A1.
Notably, all tested adaptations lead to improvements over the baseline model, indicating that integrating navigation information positively impacts the prediction performance.
However, the results suggest that the navigation loss is not a necessary component for achieving these improvements, as SceneMotion-A1 outperforms SceneMotion-A2, which incorporates the navigation loss.
On the other hand, the late fusion strategy (SceneMotion-A3) also shows improvements over the baseline but comes with a trade-off.
This approach requires additional model parameters due to the inclusion of an additional \textit{local} and  \textit{reduction decoder}, which increase the model's complexity.
Despite these additional parameters, SceneMotion-A3 does not outperform the early fusion strategy without navigation loss (SceneMotion-A1), further emphasizing the effectiveness of the latter.

In summary, the evaluation highlights that early fusion without navigation loss is the most effective and efficient strategy for integrating navigation information into the prediction model. This approach achieves the best balance between performance gains and model complexity.

\subsection{nuPlan Open-loop Challenge}
\label{sec:eval_open_loop}
Next, we show the results of the open-loop planning evaluation of the nuPlan simulator.
Unlike other approaches (e.g., \cite{dauner2023pdm}, \cite{cheng2024pluto}), we rely solely on the direct output of our model without applying any post-processing heuristics to refine or rank the predicted trajectories.
For evaluation, we directly select the trajectory with the highest confidence for the ego vehicle as provided by the model.

\begin{table}[h]
    \vspace*{1mm}
    \centering
    \caption{
        Open-loop evaluation on the Val14 split.
        Reported is the Open-Loop Score (OLS) for several baseline planners, compared to our prediction baseline SceneMotion and its adaptations A1–A3.
    }
    \label{tab:OLS}
    \begin{tabular}{lc}
        \toprule
        Method                                                 & OLS\,$\uparrow$ \\
        \midrule
        LogReplay (upper bound)                                & 100             \\
        \midrule
        IDM \cite{treiber2000idm} (as in \cite{dauner2023pdm}) & 38              \\
        PDM-Open \cite{dauner2023pdm}                          & 86              \\
        PDM-Closed \cite{dauner2023pdm}                        & 42              \\
        PDM-Hybrid \cite{dauner2023pdm}                        & 84              \\
        \midrule
        SceneMotion (Baseline) \cite{wagner2024scenemotion}    & 82.87           \\
        SceneMotion-A1                                         & 83.59           \\
        SceneMotion-A2                                         & 81.65           \\
        SceneMotion-A3                                         & 78.25           \\
        \bottomrule
    \end{tabular}
\end{table}

The SceneMotion baseline achieves an OLS of 82.87, which comparable to the PDM planners.
However, the adaptations of SceneMotion show varying degrees of improvement.
SceneMotion-A1 achieves the highest OLS among the adaptations with an OLS improvement of around $0.9\,\%$ against the baseline, indicating that early route fusion without navigation loss positively impacts open-loop planning performance.
While SceneMotion-A2 and SceneMotion-A3 -- which incorporate navigation loss -- perform slightly better than the baseline in the prediction metrics, they perform slightly worse on the OLS.

The early fusion strategy (SceneMotion-A1) remains the most effective adaptation, achieving modest improvements over the baseline without introducing additional complexity.
An analysis of the scenario specific scores reveals, that the model especially exhibits difficulties in scenarios requiring lateral planning maneuvers.

It is important to note that the nuPlan benchmark is designed such that high scores can often be achieved without requiring complex maneuvers like lane changes.
For example, the PDM methods \cite{dauner2023pdm} primarily plan in the longitudinal direction and still achieve strong results.
As a consequence, the SceneMotion model is capable of generating more complex trajectories than required by the benchmark.
This also means that, in many benchmark scenarios, the model does not need to make challenging routing decisions to follow the predefined route, and thus the impact of navigation information on overall performance is limited.

Despite this limitation, we chose the nuPlan benchmark because it is widely recognized and allows for direct comparison with other approaches.
However, there are still scenarios where navigation information proves beneficial.
As illustrated in \cref{fig:visual-evaluation}, in a left-turn intersection scenario, the inclusion of navigation information leads to more robust and traffic rule-compliant planning. Without navigation knowledge (see \cref{fig:visual-evaluation-a,fig:visual-evaluation-b,fig:visual-evaluation-c,fig:visual-evaluation-d}), the baseline model predicts a trajectory that enters an oncoming lane, violating traffic rules.
In contrast, SceneMotion-A1 (see \cref{fig:visual-evaluation-e,fig:visual-evaluation-f,fig:visual-evaluation-g,fig:visual-evaluation-h}) generates predictions that remain aligned with the intended route and demonstrate behavior consistent with the ego vehicle’s goal.

\section{Conclusions and Future Work}
\label{sec:conclusion}
In this work, we introduced different architectural strategies to integrate navigation information into a transformer-based prediction model.
We evaluated these strategies on the nuPlan dataset, demonstrating that navigation information can enhance the model's performance in predicting future trajectories.
Although the nuPlan dataset serves as a useful foundation for assessing the influence of navigation information, it may be overly simplistic to comprehensively reflect its full potential impact.
Future work should explore more complex benchmarks, such as interPlan \cite{hallgarten2024interplan}, where the benefits of incorporating navigation information are likely to be more pronounced.
Additionally, navigation information can be represented as high-level commands, as provided in NAVSIM \cite{dauner2024navsim} and utilized, for instance, by the navigation-guided end-to-end model SSR \cite{li2025navigationguidedsparsescenerepresentation}.
Using only navigation commands, rather than map-based navigation information, reduces the amount of data required by the model and may improve efficiency.
\section*{Acknowledgements}

We acknowledge the financial support for this work by the Federal Ministry of Education and Research of Germany (BMBF) within the project autotech.agil (FKZ 01IS22088T).
This work was performed on the HoreKa supercomputer funded by the Ministry of Science, Research and the Arts Baden-Württemberg and by the Federal Ministry of Education and Research.
 
\balance %
\printbibliography

\end{document}